\documentclass[10pt,twocolumn,letterpaper]{article}

\usepackage{cvpr}              

\usepackage{graphicx}
\usepackage{amsmath}
\usepackage{amssymb}
\usepackage{booktabs}

\usepackage{dashrule}

\usepackage{array}
\usepackage{colortbl}
\usepackage{arydshln}
\usepackage{multirow}
\usepackage{multicol}
\usepackage{dashrule}
\usepackage{fancyhdr}
\usepackage{bm}
\usepackage{color}
\usepackage{colortbl}
\definecolor{mygray}{gray}{.9}
\usepackage[pagebackref,breaklinks,colorlinks]{hyperref}

\def\cvprPaperID{2641} 
\def\confName{CVPR}
\def\confYear{2022}

\begin{document}
	
	\title{MSDN: Mutually Semantic Distillation Network for Zero-Shot Learning}

	\author{Shiming Chen$^{1}$, Ziming Hong$^{1}$, Guo-Sen Xie$^{2}$, Wenhan Yang$^{3}$, Qinmu Peng$^{1}$, Kai Wang$^{4}$,\\Jian Zhao$^{5}$, Xinge You$^{1\thanks{Corresponding author}}$\\
		$^{1}$Huazhong University of Science and Technology (HUST), China \\
		$^{2}$Nanjing University of Science and Technology, China \quad
		$^{3}$Nanyang Technological University, Singapore\\
		$^{4}$National University of Singapore, Singapore\quad
		$^{5}$Institute of North Electronic Equipment, China\\
		{\tt\small \{shimingchen, pengqinmu, youxg\}@hust.edu.cn \quad hoongzm@gmail.com}
	}

	\maketitle
	%
	%

	\begin{abstract}
		The key challenge of zero-shot learning (ZSL) is how to infer the latent semantic knowledge between visual and attribute features on seen classes, and thus achieving a desirable knowledge transfer to unseen classes. Prior works either simply align the global features of an image with its associated class semantic vector or utilize unidirectional attention to learn the limited latent semantic representations, which could not effectively discover the intrinsic semantic knowledge (\textit{e.g.}, attribute semantics) between visual and attribute features. To solve the above dilemma, we propose a Mutually Semantic Distillation Network (MSDN), which progressively distills the intrinsic semantic representations between visual and attribute features for ZSL. MSDN incorporates an attribute$\rightarrow$visual attention sub-net that learns attribute-based visual features, and a visual$\rightarrow$attribute attention sub-net that learns visual-based attribute features. By further introducing a semantic distillation loss, the two mutual attention sub-nets are capable of learning collaboratively and teaching each other throughout the training process. The proposed MSDN yields significant improvements over the strong baselines, leading to new state-of-the-art performances on three popular challenging benchmarks. Our  codes have been available at: \url{https://github.com/shiming-chen/MSDN}.
	\end{abstract}
	
	\section{Introduction}\label{sec1}
	Recently, deep learning performs achievements on object recognition \cite{He2016DeepRL,wang2020region,wang2020suppressing}. Based on the prior knowledge of seen classes, humans possess a remarkable ability to recognize new concepts (classes) using shared and distinct attributes of both seen and unseen classes \cite{Lampert2009LearningTD}. Inspired by this cognitive competence, zero-shot learning (ZSL) is proposed under a challenging image classification setting to mimic the human cognitive process \cite{Larochelle2008ZerodataLO,Palatucci2009ZeroshotLW}.  ZSL aims to tackle the unseen class recognition problem by transferring semantic knowledge from seen classes to unseen ones. It is usually based on the assumption that both seen and unseen classes can be described through the shared semantic descriptions (\textit{e.g.}, attributes) \cite{Lampert2014AttributeBasedCF}. Based on the classes that a model sees in the testing phase, ZSL methods can be categorized into conventional ZSL (CZSL) and generalized ZSL (GZSL) \cite{Xian2017ZeroShotLC}, where CZSL aims to predict unseen classes, while GZSL can predict both seen and unseen ones.
	
	\begin{figure}[t]
		\small
		\begin{center}
			\includegraphics[width=8.5cm,height=4.8cm]{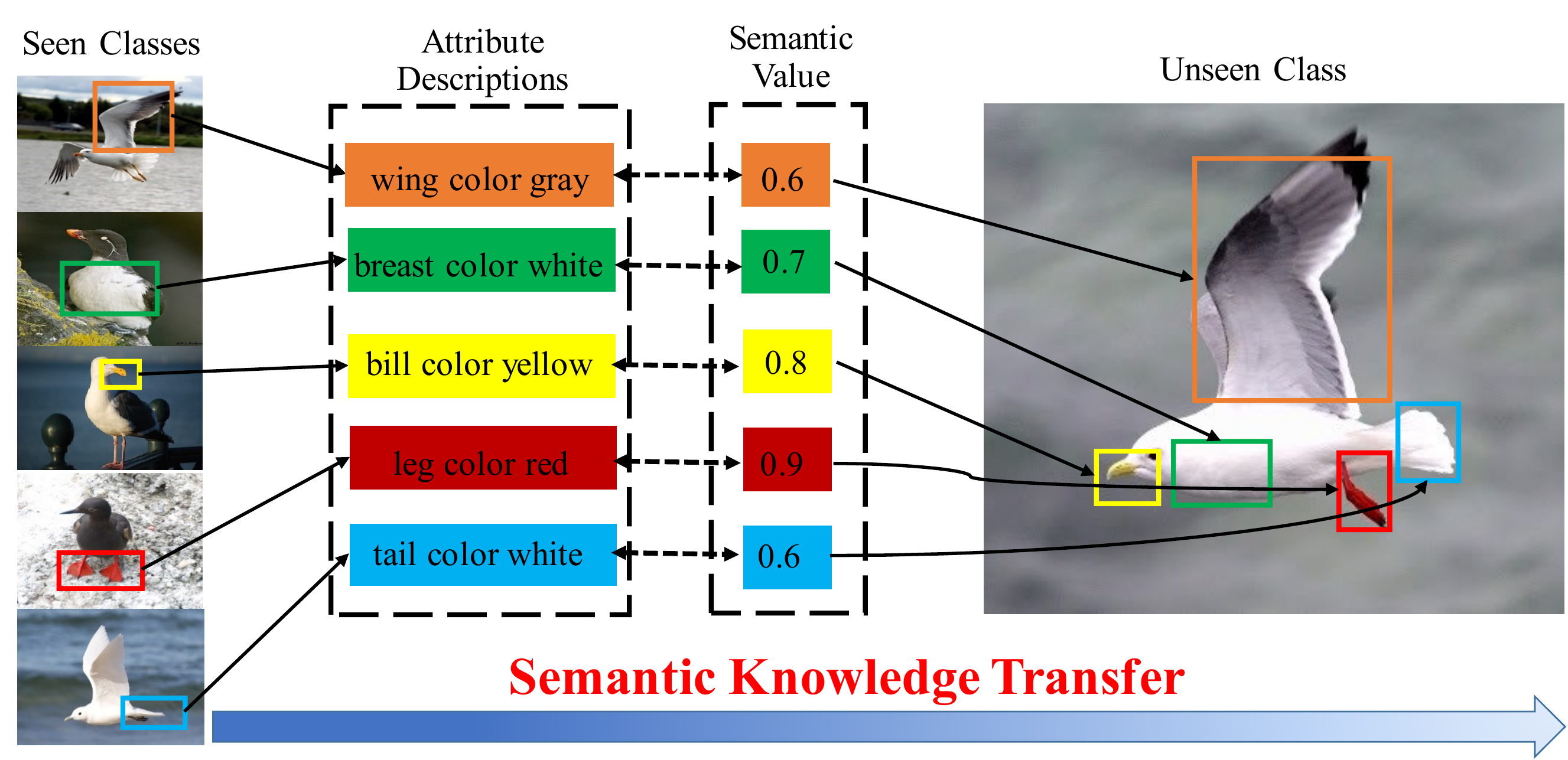}
			\vspace{-5mm}
			\caption{Motivation illustration. An unseen sample shares different partial information with a set of seen samples, and this partial information is represented as the abundant knowledge of semantic attributes (\textit{e.g.}, “bill color yellow”, “leg color red”). The key challenge of ZSL is  how to infer the latent semantic knowledge between visual and attribute features on seen classes, allowing effective knowledge transfer to unseen classes. As such, properly distilling the intrinsic semantic knowledge/representations (\textit{e.g.}, attribute semantics) between visual and attribute features from seen to unseen classes is very necessary for ZSL.} \vspace{-6mm}
			\label{fig:motivation}
		\end{center}
	\end{figure}
	
	\begin{figure*}[ht]
		\small
		\begin{center}
			\includegraphics[width=1\linewidth]{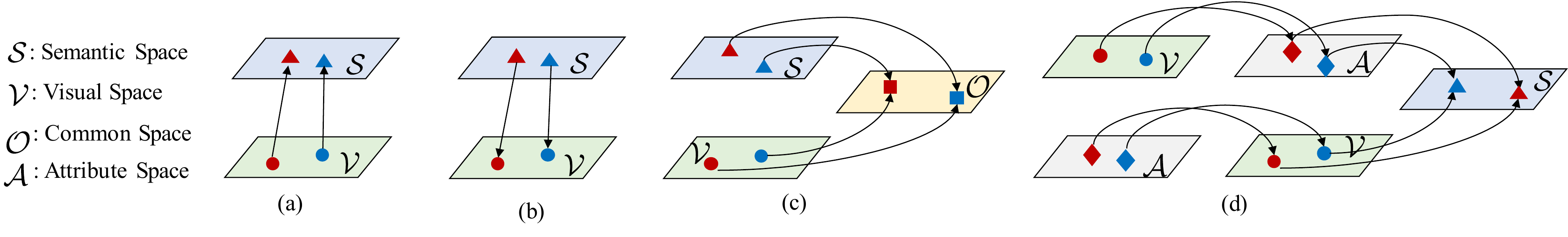}
			\\\vspace{-4mm}
			
			\caption{Four investigated ZSL paradigms. (a) Embedding-based method. (b) Generative method. (c) Common space learning-based method. (d) Ours proposed mutually semantic distillation network (MSDN). The semantic space $\mathcal{S}$ is represented by the class semantic vector annotated by humans based on the attribute descriptions. The visual space $\mathcal{V}$ is learned by a CNN backbone (\textit{e.g.}, ResNet101). The common space $\mathcal{O}$ is a shared latent space between visual mapping and semantic mapping. The attribute space $\mathcal{A}$ is learned by a language model (\textit{e.g.}, Glove \cite{Pennington2014GloveGV}). Filled triangles, circles, squares and diamonds denote the sample features in $\mathcal{S}$, $\mathcal{V}$, $\mathcal{O}$ and $\mathcal{A}$, respectively.} \vspace{-6mm}
			\label{fig:paradigm}
		\end{center}
	\end{figure*}

	ZSL has achieved significant progress, with many efforts focus on embedding-based methods, generative methods, and common space learning-based methods. As shown in Fig. \ref{fig:paradigm} {\color{red}(a)}, embedding-based methods aim to learn a visual$\rightarrow$semantic mapping to map the visual features into the semantic space for visual-semantic interaction \cite{RomeraParedes2015AnES, Akata2016LabelEmbeddingFI, Chen2018ZeroShotVR,Xie2019AttentiveRE,Xu2020AttributePN, Chen2022TransZeroAT}. The embedding-based methods usually have a large bias towards seen classes under the GZSL setting, since the embedding  function is  solely learned  by seen class samples. To solve this issue, the generative ZSL methods (see Fig. \ref{fig:paradigm}{\color{red}{\color{red}(b)}}) are proposed to learn semantic$\rightarrow$visual mapping to generate visual features of unseen classes  \cite{Arora2018GeneralizedZL,Schnfeld2019GeneralizedZA,Xian2018FeatureGN,Yu2020EpisodeBasedPG,Shen2020InvertibleZR,Vyas2020LeveragingSA,Chen2021SemanticsDF, Chen2021FREE}, and thus converting ZSL into a conventional classification problem. As shown in Fig. \ref{fig:paradigm}{\color{red}(c)}, common space learning learns a common representation space where both visual features and semantic representations are projected for knowledge transfer \cite{Frome2013DeViSEAD,Tsai2017LearningRV,Wang2017ZeroShotVR,Liu2018GeneralizedZL,Schnfeld2019GeneralizedZA, Chen2021HSVA}. However, they simply utilize the global features representations and  have neglected the fine-grained details in the training images.
	
	As shown in Fig. \ref{fig:motivation}, an unseen sample shares different partial information with a set of seen samples, and this partial information is represented as the abundant knowledge of semantic attributes (\textit{e.g.}, “bill color yellow”, “leg color red”). Thus, the key challenge of ZSL is to infer the latent semantic knowledge between visual and attribute features on seen classes, and thus allowing desirable knowledge transfer to unseen classes. Recently, some attention-based ZSL methods \cite{Xie2019AttentiveRE,Xie2020RegionGE,Zhu2019SemanticGuidedML,Xu2020AttributePN,Liu2021GoalOrientedGE,Chen2022TransZeroAT} leverage attribute descriptions as guidance to discover discriminative part/fine-grained features, enabling to match the semantic representations more accurately. Unfortunately, they simply utilize unidirectional attention, which only focuses on limited semantic alignments between visual and attribute features without any further sequential learning. As such, properly discovering the intrinsic and more sufficient semantic representations (\textit{e.g.}, attribute semantics) between visual and attribute features for knowledge transfer of ZSL is of great importance.
	
	In light of the above observation, we propose a Mutually Semantic Distillation Network (MSDN) for ZSL, as shown in Fig. \ref{fig:paradigm}{\color{red}(d)}, to explore the intrinsic semantic knowledge between visual and attribute features. MSDN consists of an attribute$\rightarrow$visual attention sub-net, which learns attribute-based visual features, and a visual$\rightarrow$attribute attention sub-net, which learns visual-based attribute features. These two mutual attention sub-nets act as a teacher-student network for guiding each other to learn collaboratively and teaching each other throughout the training process. As such, MSDN can explore the most matched attribute-based visual features and visual-based attribute features, enabling to effectively distill the intrinsic semantic representations for desirable knowledge transfer from seen to unseen classes (Fig. \ref{fig:motivation}). Specifically, each attention sub-net is trained with an attribute-based cross-entropy loss with self-calibration \cite{Zhu2019SemanticGuidedML,Huynh2020FineGrainedGZ,Xu2020AttributePN, Liu2021GoalOrientedGE,Chen2022TransZeroAT}. To encourage mutual learning between the attribute$\rightarrow$visual attention sub-net and visual$\rightarrow$attribute attention sub-net, we further introduce a semantic distillation loss that aligns each other's class posterior probabilities. The quantitative and qualitative results well demonstrate the superiority and great potential of MSDN.
	
	Our contributions are summarized as:
	\textbf{i) }We propose a Mutually Semantic Distillation Network (MSDN), orthogonal to existing ZSL methods, which distills the intrinsic semantic representations for effective knowledge transfer from seen to unseen classes for ZSL.
	\textbf{ii) }We introduce a semantic distillation loss to enable mutual learning between the attribute$\rightarrow$visual attention sub-net and visual$\rightarrow$attribute attention sub-net in MSDN, encouraging them to learn attribute-based visual features and visual-based attribute features by distilling the intrinsic semantic knowledge for semantic embedding representations.
	\textbf{iii) }We conduct extensive experiments to show that our MSDN achieves significant performance gains over the counterparts on three benchmarks, \textit{i.e.}, CUB \cite{Welinder2010CaltechUCSDB2}, SUN  \cite{Patterson2012SUNAD} and AWA2 \cite{Xian2017ZeroShotLC}.  
	%
	%
	
	\begin{figure*}[t]
		\small
		\begin{center}
			\includegraphics[width=1\linewidth]{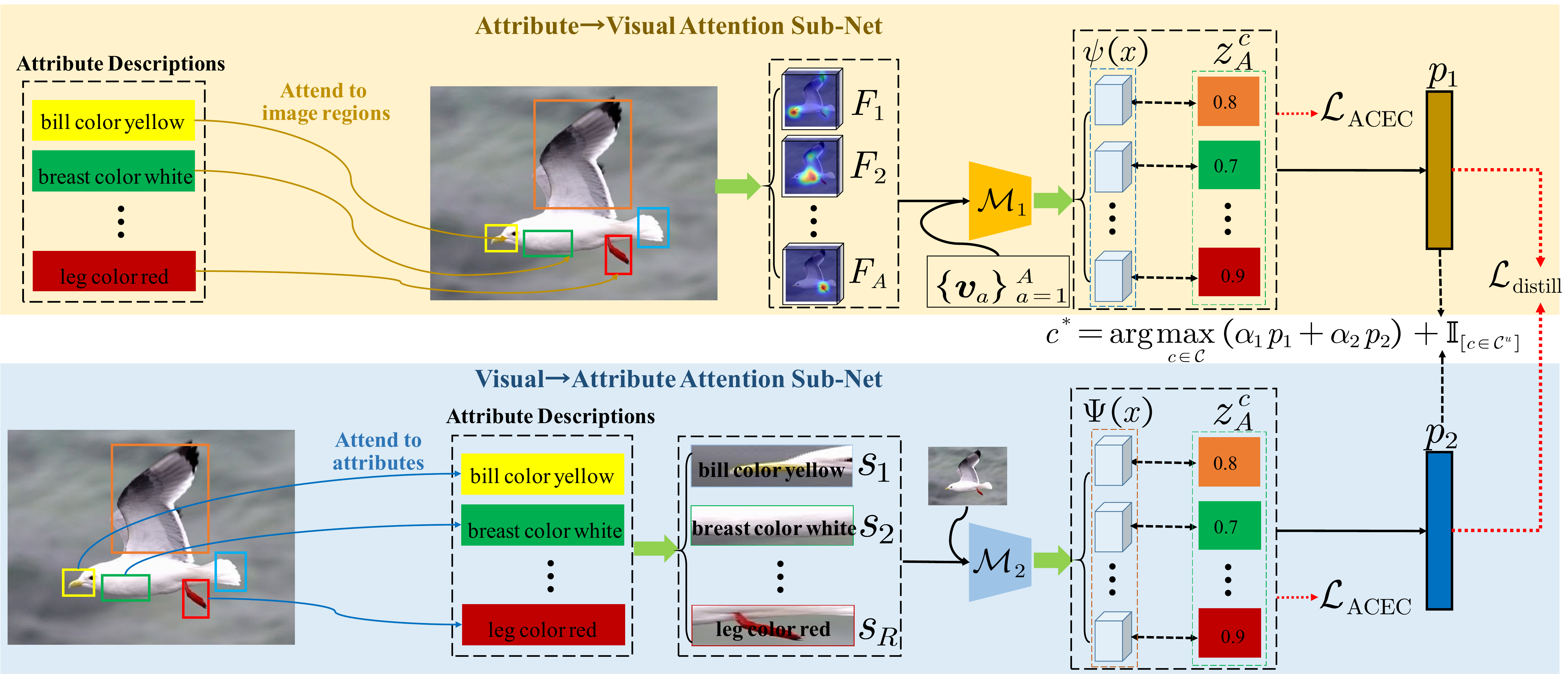}
			\\\vspace{-2mm}
			
			\caption{Illustrates of MSDN. MSDN consists of an attribute$\rightarrow$visual attention sub-net and visual$\rightarrow$attribute attention sub-net. Each sub-net is optimized with an attribute-based cross-entropy loss with self-calibration, and a semantic distillation loss to match the probability estimates of its peers for semantic distillation.} \vspace{-6mm}
			\label{fig:pipeline}
		\end{center}
	\end{figure*}
	\section{Related Work}\label{sec2}
	\subsection{Zero-Shot Learning}\label{sec2.1} To transfer semantic knowledge from seen to unseen classes, ZSL \cite{Song2018TransductiveUE,Li2018DiscriminativeLO,Xian2018FeatureGN,Xian2019FVAEGAND2AF,Yu2020EpisodeBasedPG,Min2020DomainAwareVB,Han2021ContrastiveEF,Chen2021FREE,Chou2021AdaptiveAG} learns a mapping between the visual and attribute/semantic domains. Targeting on this goal, embedding-based ZSL aims to learn a visual$\rightarrow$semantic mapping for visual-semantic interaction by mapping the visual features into the semantic space \cite{RomeraParedes2015AnES, Akata2016LabelEmbeddingFI, Chen2018ZeroShotVR,Xie2019AttentiveRE,Xu2020AttributePN}.  As the embedding is learned only on seen classes, these embedding-based methods inevitably overfit to seen classes under the GZSL setting. To mitigate this challenge, the generative ZSL models have been introduced to learn a semantic$\rightarrow$visual mapping to generate visual features of of unseen classes \cite{Arora2018GeneralizedZL,Schnfeld2019GeneralizedZA,Xian2018FeatureGN,Li2019LeveragingTI,Yu2020EpisodeBasedPG,Shen2020InvertibleZR,Vyas2020LeveragingSA, Chen2021FREE} for data augmentation. Currently, the generative ZSL usually based on variational autoencoders (VAEs) \cite{Arora2018GeneralizedZL,Schnfeld2019GeneralizedZA}, generative adversarial nets (GANs) \cite{Xian2018FeatureGN,Li2019RethinkingZL,Yu2020EpisodeBasedPG,Keshari2020GeneralizedZL,Vyas2020LeveragingSA,Chen2021FREE}, and generative flows \cite{Shen2020InvertibleZR}. Furthermore, common space learning is also employed to learns a common representation space for interaction between visual and semantic domains \cite{Frome2013DeViSEAD,Tsai2017LearningRV,Schnfeld2019GeneralizedZA, Chen2021HSVA}. However, these methods still usually yield relatively undesirable results, since they cannot capture the subtle differences between seen and unseen classes. As such, attention-based ZSL methods \cite{Xie2019AttentiveRE,Xie2020RegionGE,Zhu2019SemanticGuidedML,Xu2020AttributePN,Liu2021GoalOrientedGE,Chen2022TransZeroAT} utilize attribute descriptions as guidance to discover the more discriminative fine-grained features. They simply utilize unidirectional attention, which only focuses on limited semantic alignments between visual and attribute features without any further sequential learning. As such, properly exploring the intrinsic semantic representations between visual and attribute features for knowledge transfer of ZSL is very necessary.
	
	\subsection{Knowledge Distillation}\label{sec2.2}  To compress knowledge from a large teacher network to a small student network, knowledge distillation was proposed \cite{Hinton2015DistillingTK}. Recently, knowledge distillation is extended to optimize small deep networks starting with a powerful teacher network \cite{Romero2015FitNetsHF,Parisotto2016ActorMimicDM}. By mimicking the teacher's class probabilities and/or feature representation, distilling models convey additional information beyond the conventional supervised learning target \cite{Zhang2018DeepML,Zhai2020MultipleEB}. Motivated by these, we design a mutually semantic distillation network to learn the intrinsic semantic by semantically distilling intrinsic knowledge. The mutually semantic distillation network consists of attribute$\rightarrow$visual attention and  visual$\rightarrow$attribute attention sub-nets, which act as a teacher-student network to learn collaboratively and teach each other.

	\section{Mutually Semantic Distillation Network}\label{sec3}
	
	\noindent\textbf{Motivation.} Prior works simply i) align the global features of an image with its associated class semantic vector, neglecting the fine-grained information for knowledge transfer, or ii) utilize unidirectional attention to learn the latent semantic representations, which only focuses on limited semantic alignments between visual and attribute features without any further sequential learning. However, an unseen sample can share different partial information with a set of seen samples, and this partial information is represented as the abundant knowledge of semantic attributes, as shown in Fig. \ref{fig:motivation}. These observations prompt us to speculate that the current inferior performance of ZSL is closely related to the intrinsic semantic representations (\textit{e.g.}, attribute semantics) between visual and attribute features, which offers effective knowledge transfer to unseen classes. 
	
	To properly learn the intrinsic semantic knowledge, we propose a Mutually Semantic Distillation Network  (MSDN). Our strategy behind MSDN is to distill the intrinsic semantic knowledge from the attribute-based visual features and visual-based attribute features, which are leaned by two attention sub-nets optimized by a semantic distillation loss.  
	
	\noindent\textbf{Overview.} As illustrated in Fig. \ref{fig:pipeline}, our MSDN includes an attribute$\rightarrow$visual attention sub-net and visual$\rightarrow$attribute attention sub-net. Under the constraint of attribute-based cross-entropy loss with self-calibration, the attribute$\rightarrow$visual attention sub-net attempts to learn attribute-based visual features, and visual$\rightarrow$attribute attention sub-net aims to discover visual-based attribute representations. A semantic distillation loss encourages the two mutual attention sub-nets to learn collaboratively and teach each other throughout the training process.
	
	\noindent\textbf{Notation}. Assume that we have training data $\mathcal{D}^{s}=\left\{\left(x_{i}^{s}, y_{i}^{s}\right)\right\}$ with $C^s$ seen classes, where $x_i^s \in \mathcal{X}$ denotes the visual sample $i$, and $y_i^s \in \mathcal{Y}^s$ is the corresponding class label. Another set of unseen classes $C^u$ has unlabeled samples $\mathcal{D}^{u}=\left\{\left(x_{i}^{u}, y_{i}^{u}\right)\right\}$, where $x_{i}^{u}\in \mathcal{X}$ are the unseen class samples, and $y_{i}^{u} \in \mathcal{Y}^u$ are the corresponding labels. A set of class semantic vectors (semantic value annotated by humans according to attributes) of the class $c \in \mathcal{C}^{s} \cup \mathcal{C}^{u} = \mathcal{C}$ with $|A|$ attributes $z^{c}=\left[z_{1}^{c}, \ldots, z_{A}^{c}\right]^{\top}= \phi(y)$ which helps knowledge transfer from seen to unseen classes.  Note that we also use the semantic attribute vectors of each attribute $ A=\{a_{1}, \ldots, a_{K}\}$ learned by GloVe \cite{Pennington2014GloveGV}.

	\subsection{Attribute$\rightarrow$Visual Attention Sub-net}\label{sec3.1}
	Learning the fine-grained features for attribute localization is important in ZSL \cite{Xie2019AttentiveRE,Xie2020RegionGE,Zhu2019SemanticGuidedML,Xu2020AttributePN}. As the first component of our MSDN, we proposed an attribute$\rightarrow$visual attention sub-net to localize the most relevant image regions to the attribute to extract attribute-based visual features from a given image for each attribute.  It expects two inputs: a set of visual features of the image $ V=\{v_{1}, \ldots, v_{R}\}$, such that each visual feature encodes a region in an image; a set of semantic attribute vectors $ A=\{a_{1}, \ldots, a_{K}\}$. We can attend to image regions with respect to each attribute, and compare each attribute to the corresponding attended visual region features to determine the importance of each attribute. For the $k$-th attribute, we first define its attention weights of focusing on the $r$-th region of one image as:
	\begin{gather}
	\small
	\label{eq:attr_att}
	\beta_k^r = \frac{\exp \left(a_{k}^{\top} W_{1} v_{r}\right)}{\sum_{k=1}^{K} \exp \left(a_{k}^{\top} W_{1} v_{r}\right)},
	\end{gather}
	where $W_{1}$ is a learnable matrix to calculate the visual feature of each region and measure the similarity between each semantic attribute vector. As such, we can get a set of attention weights $\{\beta_k^r\}_{r=1}^{R}$.
	
	We then extract the attribute-based visual features for each attribute based on the attention weights. For example, we get the $k$-th attribute-based visual feature $F_k$, which is relevant to the $k$-th attribute according to the semantic vector $a_k$. It is formulated as:
	\begin{gather}
	\small
	\label{eq:v_feature}
	F_k = \sum_{r=1}^{R} \beta_k^r v_r.
	\end{gather}
	Intuitively, $F_k$ captures the visual evidence used to localize the corresponding semantic attribute in the image. If an image has an obvious attribute $a_k$, the model will assign a high positive score to the $k$-th attribute.  Otherwise, the model will assign a negative score to the $k$-th attribute. Thus, we get a set of attribute-based visual features $F=\{F_1,F_2,\cdots,F_K\}$.
	
	After extracting the attribute-based visual features, we further introduce a mapping function $\mathcal{M}_1$ to map them into the semantic embedding space. To encourage the mapping to be more accurate, we take the semantic attribute vectors $A=\{a_1, a_2,\cdots,a_K\}$ as support. Specifically, $\mathcal{M}_1$ matches the attribute-based visual feature $F_k$ with its corresponding semantic attribute vector $a_k$, formulated as:
	\begin{gather}
	\small
	\label{eq:m_1}
	\psi_k=\mathcal{M}_1(F_k)= a_{k}^{\top} W_2 F_k,
	\end{gather}
	where $W_2$ is an embedding matrix that embeds $F$ into the semantic space. Essentially, $\psi_k$ is an attribute score that represents the confidence of having the $k$-th attribute in an given image. Finally, MSDN obtains a mapped semantic embedding $\psi(x)=\{\psi_1,\psi_2, \cdots,\psi_K\}$ for each image.
	
	\subsection{Visual$\rightarrow$Attribute Attention Sub-net}\label{sec3.2}
	Analogously, we design a visual$\rightarrow$attribute attention sub-net to learn visual-based semantic attribute representations. They are complementary to the attribute-based visual features, enabling them to calibrate each other to discover the intrinsic semantic representations between visual and attribute features. We can first attend to semantic attributes with respect to each image region. Formally, we define its attention weights to focus on the $k$-th attribute as:
	\begin{gather}
	\small
	\label{eq:vis_att}
	\tau_r^k = \frac{\exp \left(v_{r}^{\top} W_{3} a_{k}\right)}{\sum_{r=1}^{R} \exp \left(v_{r}^{\top} W_{3} a_{k}\right)},
	\end{gather}
	where $W_{3}$ is a learnable matrix to measure the similarity between the semantic attribute vector and each visual region feature. Thus, we can get a set of attention weights $\{\tau_r^k\}_{k=1}^{K}$, which is used to extract visual-based attribute features. It is formulated as:
	\begin{gather}
	\small
	\label{eq:a_feature}
	S_r = \sum_{k=1}^{K} \tau_r^k a_k.
	\end{gather}
	Intrinsically, $S_r$ is the visual semantic representations, which is aligned to the $F_k$. We further introduce another mapping function $\mathcal{M}_2$ to map these visual-based attribute features $S=\{S_1,S_2,\cdots,S_R\}$ into semantic space:
	\begin{gather}
	\small
	\label{eq:m_2}
    \bar{\Psi}_r=\mathcal{M}_2(S_r)= v_{r}^{\top} W_4 S_r,
	\end{gather}
	where $W_4$ is an embedding matrix. Given a set of $V=\{v_{1}, \ldots, v_{R}\}$, MSDN gets the mapped semantic embedding $\bar{\Psi}(x)=\{\bar{\Psi}_1,\bar{\Psi}_2, \cdots,\bar{\Psi}_R\}$ for the attributes of one image. To enable the learned semantic embedding $\bar{\Psi}(x_i)$ is $R$-$dim$ to match with the dimension of class semantic vector ($K$-$dim$), it is further mapped into semantic attribute space with $K$-$dim$, formulated as $\Psi(x_i)=\bar{\Psi}(x_i)\times Att = \bar{\Psi}(x_i) \times(V^{\top}W_{att}A)$, where $W_{att}$ is a learnable matrix. 
	\subsection{Model Optimization}\label{sec3.3}
	To optimize MSDN, each attention sub-net is trained with a supervised learning loss, \textit{i.e.}, attribute-based cross-entropy loss with self-calibration. To encourage mutual learning between the two attention sub-nets, we introduce a semantic distillation loss that aligns each other's class posterior probabilities.

	\noindent\textbf{Attribute-Based Cross-Entropy Loss.}
	Since the associated image and attribute embeddings are projected near their class semantic vector $z^{c}$ when an attribute is visually present in an image, we take the attribute-based cross-entropy loss with self-calibration \cite{Zhu2019SemanticGuidedML,Huynh2020FineGrainedGZ,Xu2020AttributePN} (denoted as $\mathcal{L}_{\text{ACEC}}$) to optimize the parameters of the MSDN. This enables the image to have the highest compatibility score with its corresponding class semantic vector. Given a batch of $n_b$ training images $\{x_i^s\}_{i=1}^{n_b}$ with their corresponding class semantic vectors $z^c$,  $\mathcal{L}_{\text{ACEC}}$ is defined as:
	\begin{gather}
	\small
	\label{eq:L_ACEC}\
	\begin{aligned}
	\mathcal{L}_{\text{ACEC}}&=-\frac{1}{n_{b}} \sum_{i=1}^{n_{b}} [\log \frac{\exp \left(f(x_i) \times z^{c}\right)}{\sum_{\hat{c} \in \mathcal{C}^s} \exp \left(f(x_i)\times z^{\hat{c}} \right)}\\
	&-\lambda_{\text{cal}}\sum_{c^{\prime=1}}^{\mathcal{C}^u} \log \frac{\exp \left(f(x_i) \times z^{c^{\prime}} + \mathbb {I}_{\left[c^{\prime}\in\mathcal{C}^u\right]}\right)}{\sum_{\hat{c} \in \mathcal{C}} \exp \left(f(x_i) \times z^{\hat{c}} + \mathbb {I}_{\left[\hat{c}\in\mathcal{C}^u\right]}\right)}],
	\end{aligned}
	\end{gather}
	where $f(x_i)= \psi(x_i)$ for attribute$\rightarrow$visual attention sub-net and $f(x_i)=\Psi(x_i)$ for visual$\rightarrow$semantic attention sub-net, $\mathbb {I}_{\left[c\in \mathcal{C}^u\right]}$ is an indicator function (\textit{i.e.}, it is 1 when $c\in\mathcal{C}^u$, otherwise -1), and $\lambda_{\text{cal}}$ is a weight to constrol the self-calibration term. Intuitively, $\mathcal{L}_{\text{ACEC}}$ encourages non-zero probabilities to be assigned to the unseen classes during training, thus MSDN produces a large probability for the true unseen class when given test unseen samples.

	\noindent \textbf{Semantic Distillation Loss.}
	To enable the two mutual attention sub-nets to learn collaboratively and teach each other throughout the training process, we further introduce a semantic distillation loss $\mathcal{L}_{\text{distill}}$ for optimization. $\mathcal{L}_{\text{distill}}$ consists of a Jensen-Shannon Divergence (JSD) and an $\ell_2$ distance between the predictions of the two attention sub-nets (\textit{i.e.}, $p_1=\{\psi(x_i)\times z^1, \cdots, \psi(x_i)\times z^C\}$ and $p_2=\{\Psi(x_i)\times z^1, \cdots, \Psi(x_i)\times z^C\}$), formulated as:
	\begin{gather}
	\small
	\begin{aligned}
	\label{eq:L_{distill}}
	\mathcal{L}_{\text{distill}}&=\frac{1}{n_{b}} \sum_{i=1}^{n_{b}}[\underbrace{\frac{1}{2}\left(D_{K L}\left(p_1(x_i) \| p_2(x_i)\right)+D_{K L}\left(p_2(x_i) \| p_1(x_i)\right)\right)}_{\text{JSD}} \\
	&+  \underbrace{\|p_1(x_i)-p_2(x_i)\|_{2}^{2}}_{\ell_2}],
	\end{aligned}
	\end{gather}
	where
	\begin{gather}
	\small
	\label{eq:L_{distill1}}
	D_{KL}(p||q)=\sum_{c=1}^{C^s}p^c \log(\frac{p^c}{q^c}).
	\end{gather}

	\noindent\textbf{Overall Loss.} Finally, we define the overall loss function of MSDN as:
	\begin{gather}
	\small
	\label{Eq:L_final}
	\mathcal{L}_{\text{total}}=  \mathcal{L}_{\text{ACEC}} + \lambda_{\text{distill}}\mathcal{L}_{\text{distill}},
	\end{gather}
	where $\lambda_{\text{distill}}$ is a weight to control the semantic distillation loss.
	
	\subsection{Zero-Shot Prediction}\label{sec3.4}
	After training MSDN, We first obtain the embedding features of a test
	instance $x_i$ in the semantic space w.r.t. the semantic$\rightarrow$visual and visual$\rightarrow$semantic attention sub-nets, \textit{i.e.}, $\psi(x)$ and $\Psi(x)$. Then, We fuse their predictions using two combination coefficients $(\alpha_1, \alpha_2)$ to
	predict the test label of $x_i$ with an explicit calibration, formulated as:
	\begin{gather}
	\small
	\label{Eq:prediction}
	c^{*}=\arg \max _{c \in \mathcal{C}^u/\mathcal{C}}(\alpha_1\psi(x_i)+\alpha_2\Psi(x_i))^{\top} \times z^{c}+\mathbb {I}_{\left[c\in\mathcal{C}^u\right]}.
	\end{gather}
	Here, $\mathcal{C}^u$/$\mathcal{C}$ corresponds to the CZSL/GZSL setting.

		\begin{table*}[ht]
		\small
		\centering  
		\caption{Results ~($\%$) of the state-of-the-art CZSL and GZSL modes on CUB, SUN and AWA2, including generative methods, common space-based methods, and embedding-based methods. The best and second-best results are marked in \textbf{\color{red}Red} and \textbf{\color{blue}Blue}, respectively. The symbol “--” indicates no results. The symbol “{\color{red}*}” denotes attention-based methods.}\label{Table:SOTA}\vspace{-3mm}
		\resizebox{1.0\linewidth}{!}{\small
			\begin{tabular}{r|c|ccc|c|ccc|c|ccc}
				\hline
				\multirow{3}{*}{\textbf{Methods}} 
				&\multicolumn{4}{c|}{\textbf{CUB}}&\multicolumn{4}{c|}{\textbf{SUN}}&\multicolumn{4}{c}{\textbf{AWA2}}\\
				\cline{2-5}\cline{6-9}\cline{9-13}
				&\multicolumn{1}{c|}{CZSL}&\multicolumn{3}{c|}{GZSL}&\multicolumn{1}{c|}{CZSL}&\multicolumn{3}{c|}{GZSL}&\multicolumn{1}{c|}{CZSL}&\multicolumn{3}{c}{GZSL}\\
				\cline{2-5}\cline{6-9}\cline{9-13}
				\textbf{} 
				&\rm{acc}&\rm{U} & \rm{S} & \rm{H} &\rm{acc}&\rm{U} & \rm{S} & \rm{H} &\rm{acc}&\rm{U} & \rm{S} & \rm{H} \\
				
				\hline
				\textbf{Generative Methods} &&&&&&&&&&&&\\ 
				\rowcolor{mygray}f-CLSWGAN(CVPR'18)~\cite{Xian2018FeatureGN}    &57.3&43.7&57.7& 49.7&60.8&42.6&36.6&39.4&68.2&57.9&61.4&59.6\\
				f-VAEGAN-D2(CVPR'19)~\cite{Xian2019FVAEGAND2AF}&61.0&48.4&60.1& 53.6&\textbf{\color{blue}64.7}&45.1&38.0&41.3&71.1&57.6&70.6&63.5\\
				\rowcolor{mygray}E-PGN(CVPR'20)~\cite{Yu2020EpisodeBasedPG}&\textbf{\color{blue}72.4}&52.0&61.1&56.2&--&--&--&--&\textbf{\color{red}73.4}&52.6&83.5&64.6\\
				Composer$^{\color{red}*}$(NeurIPS'20)~\cite{Huynh2020CompositionalZL}&69.4&56.4&63.8&59.9&62.6& \textbf{\color{red}55.1}&22.0& 31.4&\textbf{\color{blue}71.5}& \textbf{\color{red}62.1}&77.3&\textbf{\color{red}68.8}\\
				\rowcolor{mygray}GCM-CF(CVPR'21)~\cite{Yue2021CounterfactualZA}&--&61.0&59.7&60.3&--& 47.9&37.8& \textbf{\color{blue}42.2}&--& 60.4&75.1&67.0\\
				FREE(ICCV'21)~\cite{Chen2021FREE}&--&55.7&59.9&57.7&--& 47.4&37.2& 41.7&--& 60.4&75.4&67.1\\
				\hline
				\textbf{Common Space Learning} &&&&&&&&&&&&\\ 
				\rowcolor{mygray}DeViSE(NeurIPS'13)~\cite{Frome2013DeViSEAD}&52.0&23.8&53.0&32.8&56.5&16.9&27.4&20.9&54.2&17.1&74.7&27.8\\
				DCN(NeurIPS'18)~\cite{Liu2018GeneralizedZL}&56.2&28.4&60.7&38.7&61.8&25.5&37.0&30.2&65.2&25.5&84.2&39.1\\
				\rowcolor{mygray}CADA-VAE(CVPR'19)~\cite{Schnfeld2019GeneralizedZA}&59.8&51.6&53.5&52.4&61.7&47.2&35.7&40.6&63.0&55.8&75.0&63.9\\
				SGAL(NeurIPS'19)~\cite{Yu2019ZeroshotLV}  &--& 40.9 & 55.3 & 47.0 &--& 35.5 & 34.4 & 34.9& --& 52.5 & 86.3 & 65.3\\
				\rowcolor{mygray}HSVA(NeurIPS'21)~\cite{Chen2021HSVA}&62.8&52.7&58.3&55.3&63.8&48.6&\textbf{\color{blue}39.0}&\textbf{\color{red}43.3}&--&59.3&76.6&66.8\\
				\hline 
				\textbf{Embedding-based Methods} &&&&&&&&&&&&\\   
				\rowcolor{mygray}SP-AEN(CVPR'18)~~\cite{Chen2018ZeroShotVR}      &55.4&34.7&70.6&46.6 &59.2&24.9&38.6&30.3&58.5&23.3&90.9&37.1 \\
				SGMA$^{\color{red}*}$(NeurIPS'19)~\cite{Zhu2019SemanticGuidedML} &71.0&36.7&71.3&48.5&--&--&--&--&68.8&37.6&87.1&52.5\\
				\rowcolor{mygray}AREN$^{\color{red}*}$(CVPR'19)~\cite{Xie2019AttentiveRE}&71.8&38.9&\textbf{\color{blue}78.7}&52.1&60.6&19.0&38.8&25.5&67.9&15.6&\textbf{\color{blue}92.9}&26.7 \\
				LFGAA$^{\color{red}*}$(ICCV'19)~\cite{Liu2019AttributeAF}&67.6&36.2&\textbf{\color{red}80.9}&50.0&61.5&18.5&\textbf{\color{red}40.0}&25.3&68.1&27.0&\textbf{\color{red}93.4}&41.9\\
				\rowcolor{mygray}DAZLE$^{\color{red}*}$(CVPR'20)~\cite{Huynh2020FineGrainedGZ}&66.0&56.7&59.6&58.1&59.4&\textbf{\color{blue}52.3}&24.3&33.2&67.9&60.3&75.7&67.1\\
				APN$^{\color{red}*}$(NeurIPS'20)~\cite{Xu2020AttributePN}&72.0&\textbf{\color{blue}65.3}& 69.3&\textbf{\color{blue}67.2}&61.6& 41.9&34.0&37.6&68.4&57.1&72.4&63.9\\

				\cdashline{1-13}[4pt/1pt]
				{\textbf{MSDN}}{~\textbf{(Ours)}}    &\textbf{\color{red}76.1}&\textbf{\color{red}68.7}&67.5&\textbf{\color{red}{68.1}}&\textbf{\color{red}65.8}&52.2&34.2&41.3&70.1&\textbf{\color{blue}62.0}&74.5&\textbf{\color{blue}67.7}\\
				\hline	
		\end{tabular} }\vspace{-5mm}
		\label{table:sota} 
	\end{table*}
	
	\begin{table}[ht]
		\small
		\centering
		\caption{ Ablation studies for different components of MSDN. The \textit{baseline} is the visual feature extracted from CNN backbone with a global average pooling and then mapped into semantic embedding for ZSL. The V$\rightarrow$A and A$\rightarrow$V denote visual$\rightarrow$attribute and attribute$\rightarrow$visual attention sub-nets, respectively.} \label{table:ablation}\vspace{-3mm}
		\resizebox{0.49\textwidth}{!}
		{
			\begin{tabular}{l|cc|cc}
				
				\hline
				\multirow{2}*{Method} &\multicolumn{2}{c|}{\textbf{CUB}} &\multicolumn{2}{c}{\textbf{SUN}}\\
				\cline{2-3}\cline{4-5}
				&\rm{acc}&\rm{H} &\rm{acc}& \rm{H}\\
				\hline
				\rowcolor{mygray}baseline                                                           & 57.4&49.1	& 54.8&30.5\\
				MSDN(V$\rightarrow$A) w/o $\mathcal{L}_{\text{distill}}$                  & 66.0 &55.4	& 59.2&33.8\\
				\rowcolor{mygray}MSDN(A$\rightarrow$V) w/o $\mathcal{L}_{\text{distill}}$                  & 73.4& 65.4 & 63.8&38.5\\
				MSDN(V$\rightarrow$A) w/ $\mathcal{L}_{\text{distill}}$                   & 67.9&60.8	& 62.1& 38.6\\
				\rowcolor{mygray}MSDN(A$\rightarrow$V) w/ $\mathcal{L}_{\text{distill}}$                   & 75.2&67.5 & 63.0&38.7	\\
				MSDN w/ $\mathcal{L}_{\text{distill}}$(JSD)                                    & 74.3&67.0 & 64.7&39.4	\\
				\rowcolor{mygray}MSDN w/ $\mathcal{L}_{\text{distill}}$($\ell_2$)                                    & 74.4&67.6 & 64.9 &40.8	\\
				MSDN (full)                                      &\textbf{76.1}&\textbf{68.1}&\textbf{65.8}&\textbf{41.3}\\
				\hline
			\end{tabular}
		}\vspace{-7mm}
	\end{table}

	\section{Experiments}\label{sec4}
	\noindent\textbf{Datasets.}  We evaluate our method on three challenging benchmark datasets, \textit{i.e.}, CUB (Caltech UCSD Birds 200) \cite{Welinder2010CaltechUCSDB2}, SUN (SUN Attribute) \cite{Patterson2012SUNAD} and AWA2 (Animals with Attributes 2)\cite{Xian2017ZeroShotLC}. Among these, CUB and SUN are fine-grained datasets, whereas AWA2 is a coarse-grained dataset. Following \cite{Xian2017ZeroShotLC}, we use the same seen/unseen splits and class embeddings. Specifically, CUB includes 11,788 images of 200 bird classes (seen/unseen classes = 150/50) with 312 attributes. SUN has 14,340 images from 717 scene classes (seen/unseen classes = 645/72) with 102 attributes. AWA2 consists of 37,322 images from 50 animal classes (seen/unseen classes = 40/10) with 85 attributes.

	\noindent\textbf{Evaluation Protocols.} We evaluate the top-1 accuracy on unseen classes in the CZSL setting, denoted as $acc$. For GZSL setting, we evaluate the top-1 accuracies both on seen and unseen classes (\textit{i.e.}, $S$ and $U$), respectively. Furthermore, their harmonic mean (defined as $H =(2 \times S \times U) /(S+U)$) is also employed for evaluating the performance in the GZSL setting.

	\noindent\textbf{Implementation Details.} We take a ResNet101 \cite{He2016DeepRL} pre-trained on ImageNet as the CNN backbone to extract the feature map for each image without fine-tuning. We use the RMSProp optimizer with hyperparameters (momentum = 0.9, weight decay = 0.0001) to optimize our model. We set the learning rate and batch size to 0.0001 and 50, respectively. We empirically set the loss weights $\{\lambda_{\text{cal}},\lambda_{\text{distill}}\}$ to $\{0.1,0.001\}$ for CUB and AWA2, and $\{0.0,0.01\}$ for SUN.

	\subsection{Comparision with State-of-the-Arts}\label{sec4.1}
	
	\noindent\textbf{Conventional Zero-Shot Learning.} 
	We first compare our MSDN with the state-of-the-art methods in the CZSL setting. Table \ref{table:sota} presents the results of CZSL on various datasets. Our MSDN achieves the best accuracies of 76.1\% and 65.8\% on CUB and SUN, respectively. This shows that MSDN distills the intrinsic semantic representations for distinguishing fine-grained unseen classes. As shown in Fig. \ref{fig:motivation}, MSDN can distill the semantic attributes of “bill color yellow”, “breast color white” and “leg color red” from \textit{Alifornia Gull}, \textit{Parakeet Auklet}, \textit{Pigeon Guillemot}, transferring to unseen classes (\textit{e.g.}, \textit{Red Legged Kittiwake}). As for the coarse-grained dataset (\textit{i.e.}, AWA2), MSDN still obtains competitive performance, with a top-1 accuracy of 70.1\%. Compared to other embedding-based methods, MSDN achieves new state-of-the-art on all datasets.
	
	\begin{figure*}[t]
		\begin{center}
			\hspace{0.5mm}\rotatebox{90}{\hspace{0.4cm}{\footnotesize (a) A$\rightarrow$V }}\hspace{0mm}
			\includegraphics[width=0.95\linewidth]{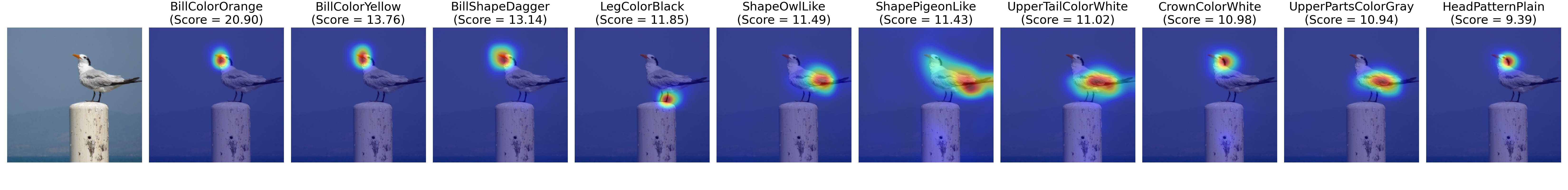}\\ 
			\hspace{0.5mm}\rotatebox{90}{\hspace{0.4cm}{\footnotesize (b) V$\rightarrow$A }}\hspace{0mm}
			\includegraphics[width=0.95\linewidth]{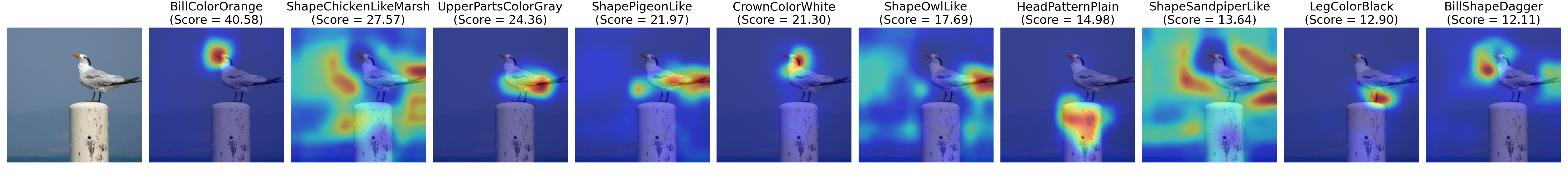}\\\vspace{-4mm}
			\caption{Visualization of attention maps for the two mutual attention sub-nets (i.e, MSDN(A$\rightarrow$V) and MSDN(V$\rightarrow$A)). The scores are the attribute scores.. (Best viewed in color)}\vspace{-6mm}
			\label{fig:att}
		\end{center}
	\end{figure*}
	
	\begin{figure*}[t]
		\begin{center}
			\hspace{0.5mm}\rotatebox{90}{\hspace{0.8cm}{\footnotesize (a) Seen Classes }}\hspace{-0.3mm}
			\includegraphics[width=14cm,height=3.5cm]{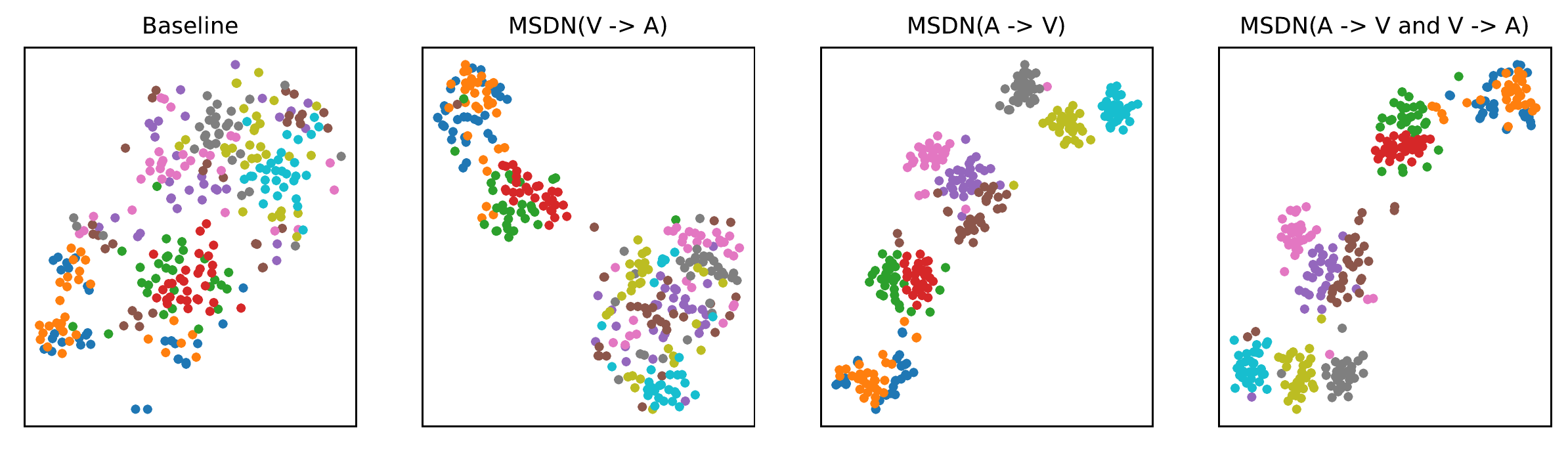}\\ 
			\hspace{0.5mm}\rotatebox{90}{\hspace{0.6cm}{\footnotesize (b) Unseen Classes }}\hspace{-0.3mm}
			\includegraphics[width=14cm,height=3.5cm]{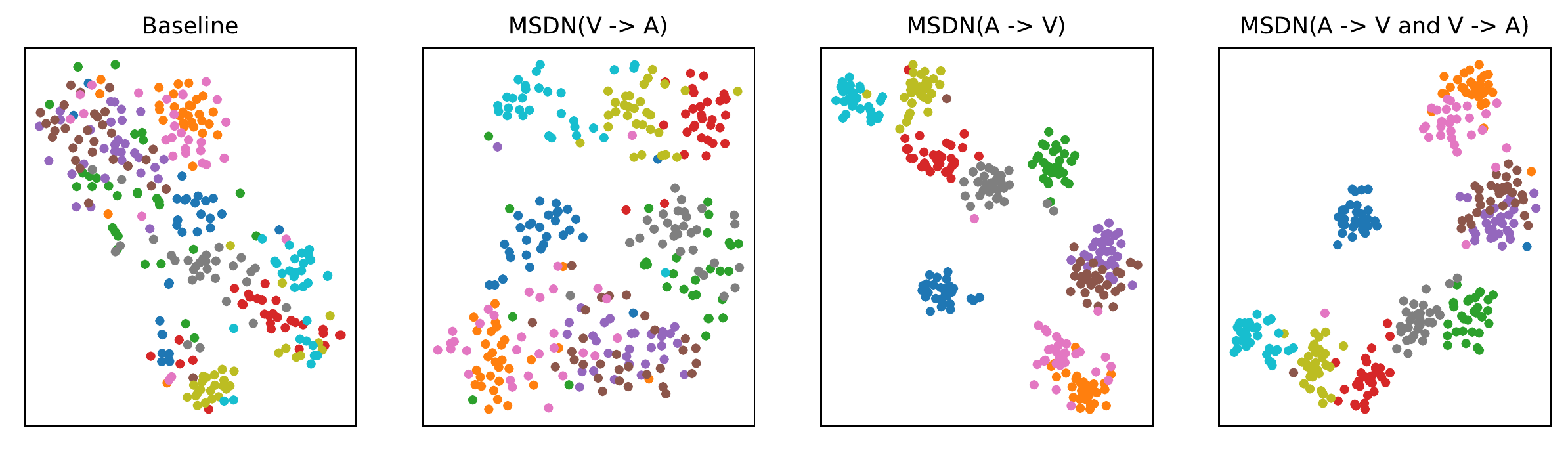}\\\vspace{-4mm}
			\caption{t-SNE visualizations of visual features for (a) seen classes and (b) unseen classes, learned by the baseline, MSDN(V$\rightarrow$A), MSDN(A$\rightarrow$V), and  MSDN(V$\rightarrow$A and A$\rightarrow$V). The 10 colors denote 10 different seen/unseen classes randomly selected from CUB.  (Best viewed in color)}\vspace{-9mm}
			\label{fig:tsne}
		\end{center}
	\end{figure*}
	
	\noindent\textbf{Generalized Zero-Shot Learning.}
	Table \ref{table:sota} also shows the results of different methods in the GZSL setting, i.e, embedding-based methods, generative methods, and common space learning methods. Interestingly, most state-of-the-art methods achieve good results on seen classes but fail on unseen classes on CUB and AWA2, while our MSDN generalizes well to unseen classes with high seen and unseen accuracies. As such, MSDN achieves good results of Harmonic mean, \textit{e.g.}, 68.1\% and 67.7\% on CUB and AWA2, respectively. These benefits come from the semantic distillation of MSDN, enabling to discover the intrinsic semantic representations for effective knowledge transfer from seen to unseen classes. Compared to attention-based ZSL methods \cite{Xie2019AttentiveRE,Xie2020RegionGE,Liu2019AttributeAF,Zhu2019SemanticGuidedML,Xu2020AttributePN} that utilize attribute descriptions as guidance to discover the more discriminative fine-grained features, our MSDN also achieves significant improvement on Harmonic mean at least 3.7\% on SUN. This demonstrates the superiority and potential of the proposed MSDN for ZSL.

	\begin{figure*}[ht]
		\begin{center}
			\hspace{-3mm}\includegraphics[width=8.7cm,height=4.1cm]{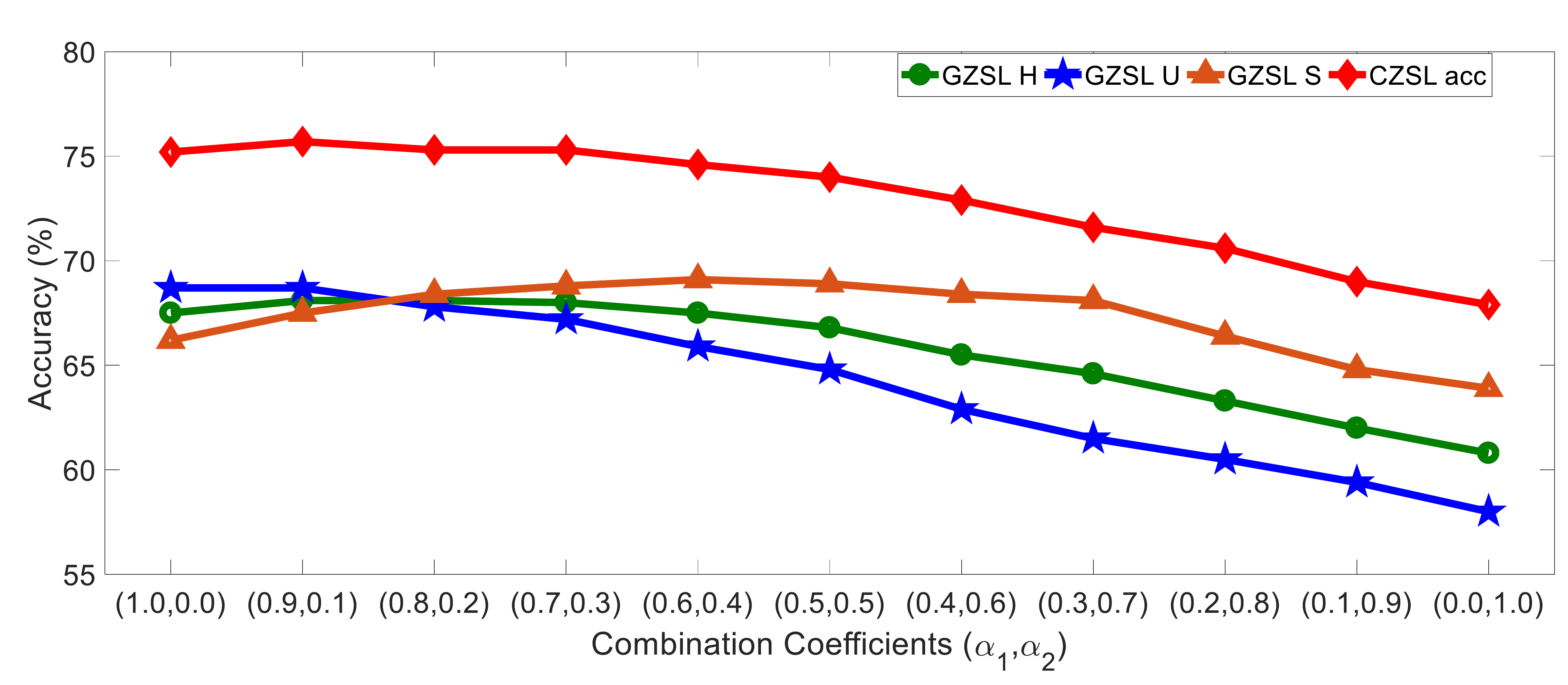}
			\includegraphics[width=8.7cm,height=4.1cm]{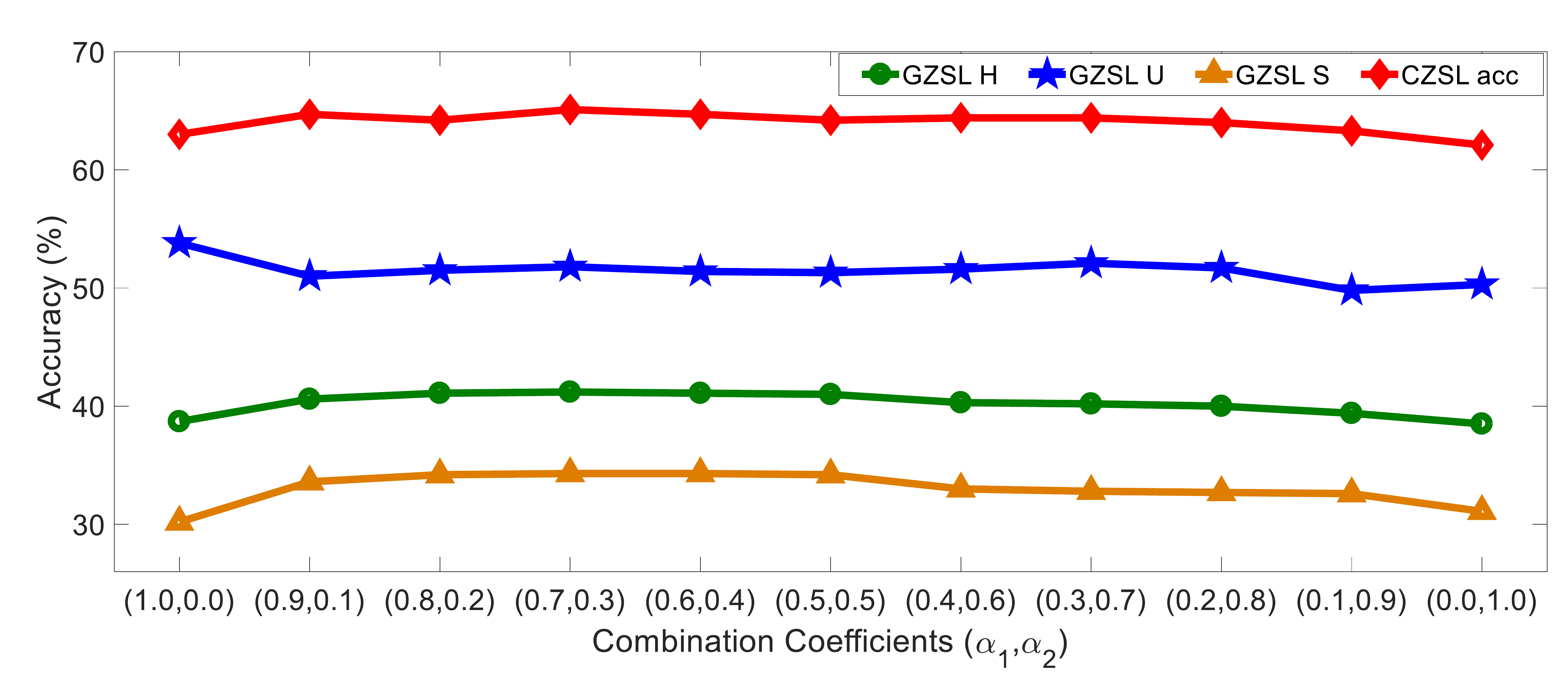}\\ \vspace{-1mm}
			(a) CUB \hspace{8cm} (b) SUN \\\vspace{-3mm}
			\caption{The effectiveness of the combination coefficients ($\alpha_{1},\alpha_{2}$) between the attribute$\rightarrow$visual and visual$\rightarrow$attribute attention sub-nets.}\vspace{-8mm}
			\label{fig:combination}
		\end{center}
	\end{figure*}

	\begin{figure*}[ht]
		\begin{center}
			\hspace{-5mm}\includegraphics[width=4.8cm,height=3.45cm]{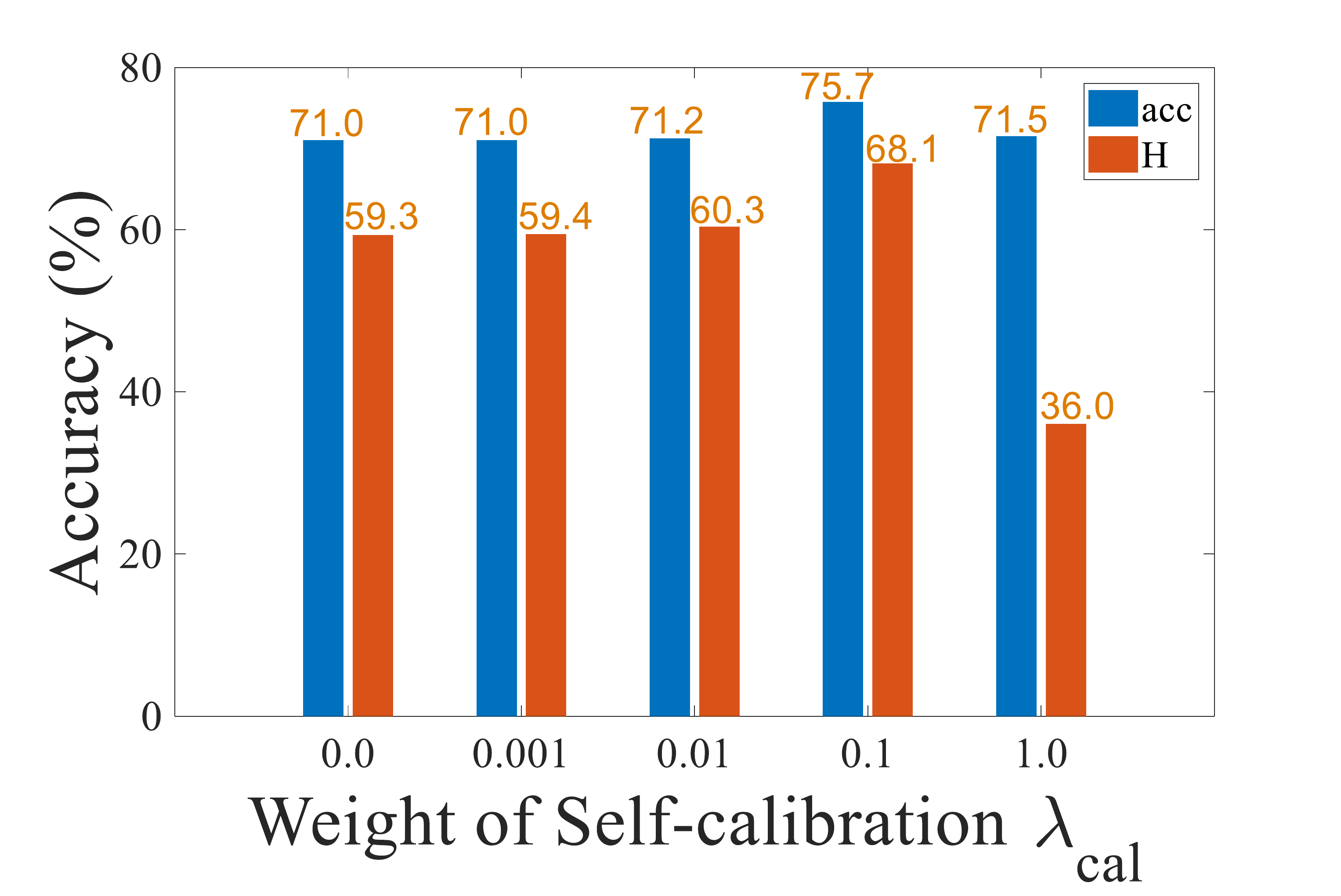}\hspace{-5mm}
			\includegraphics[width=4.8cm,height=3.45cm]{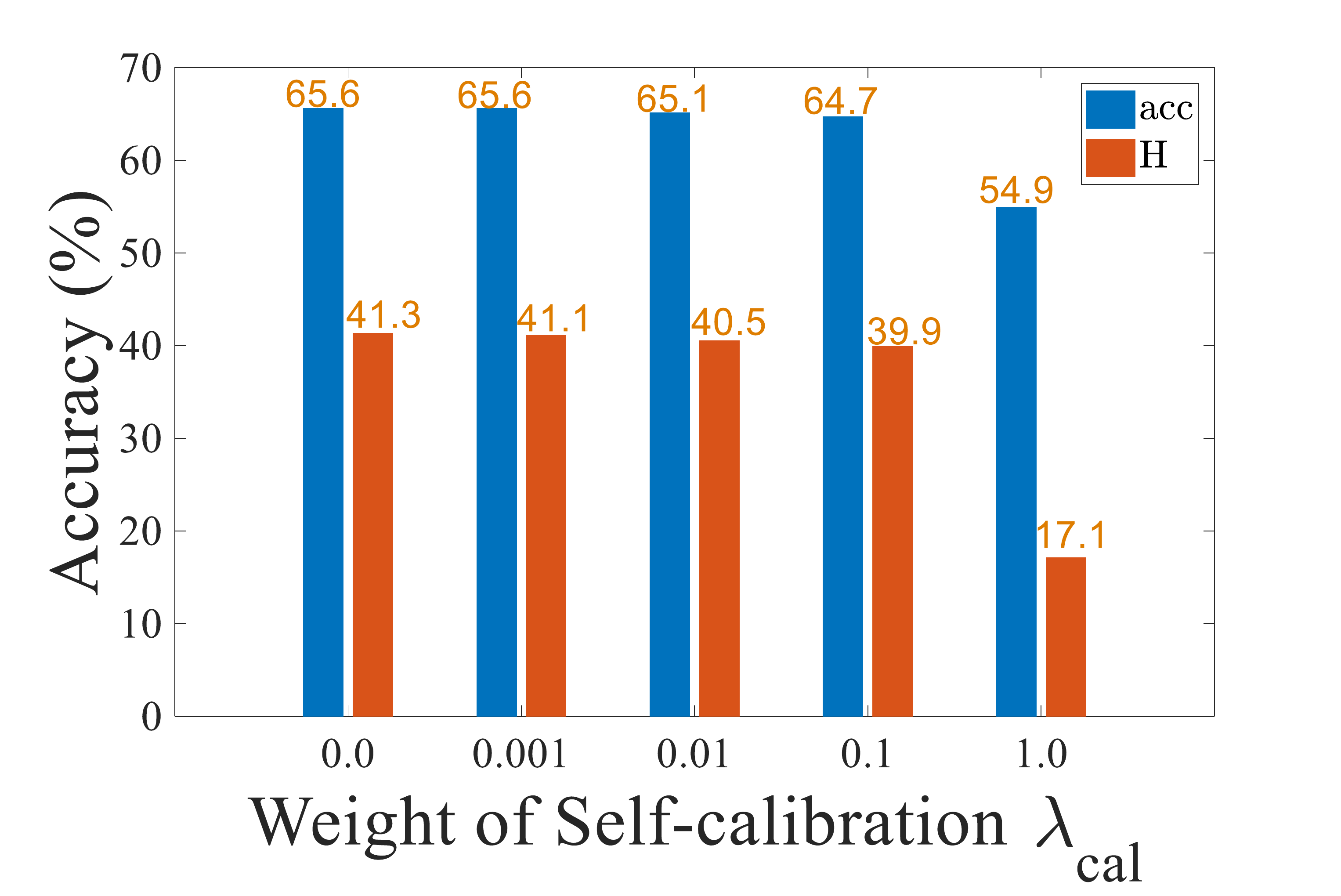}\hspace{-5mm}
			\includegraphics[width=4.8cm,height=3.45cm]{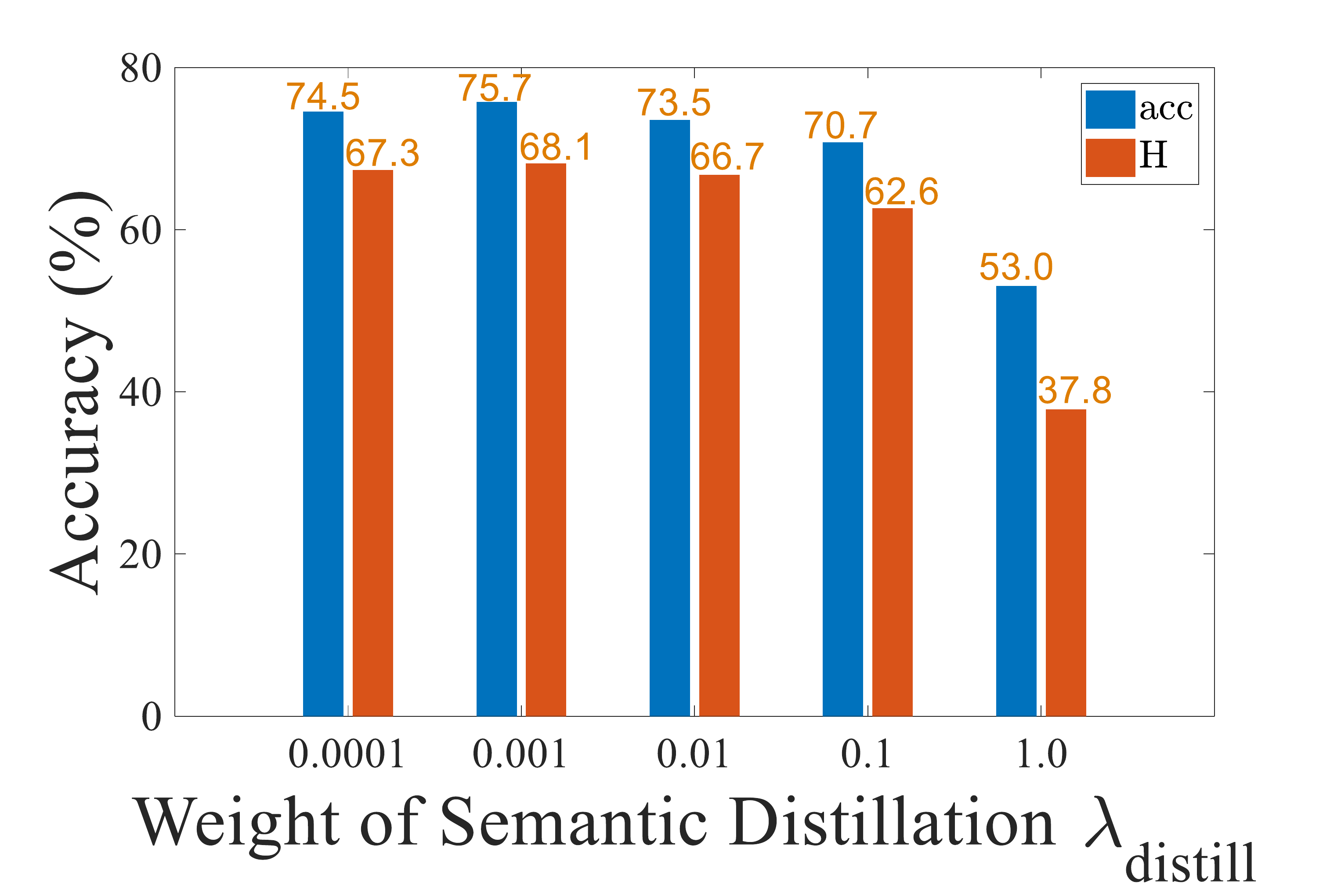}\hspace{-5mm}
			\includegraphics[width=4.8cm,height=3.45cm]{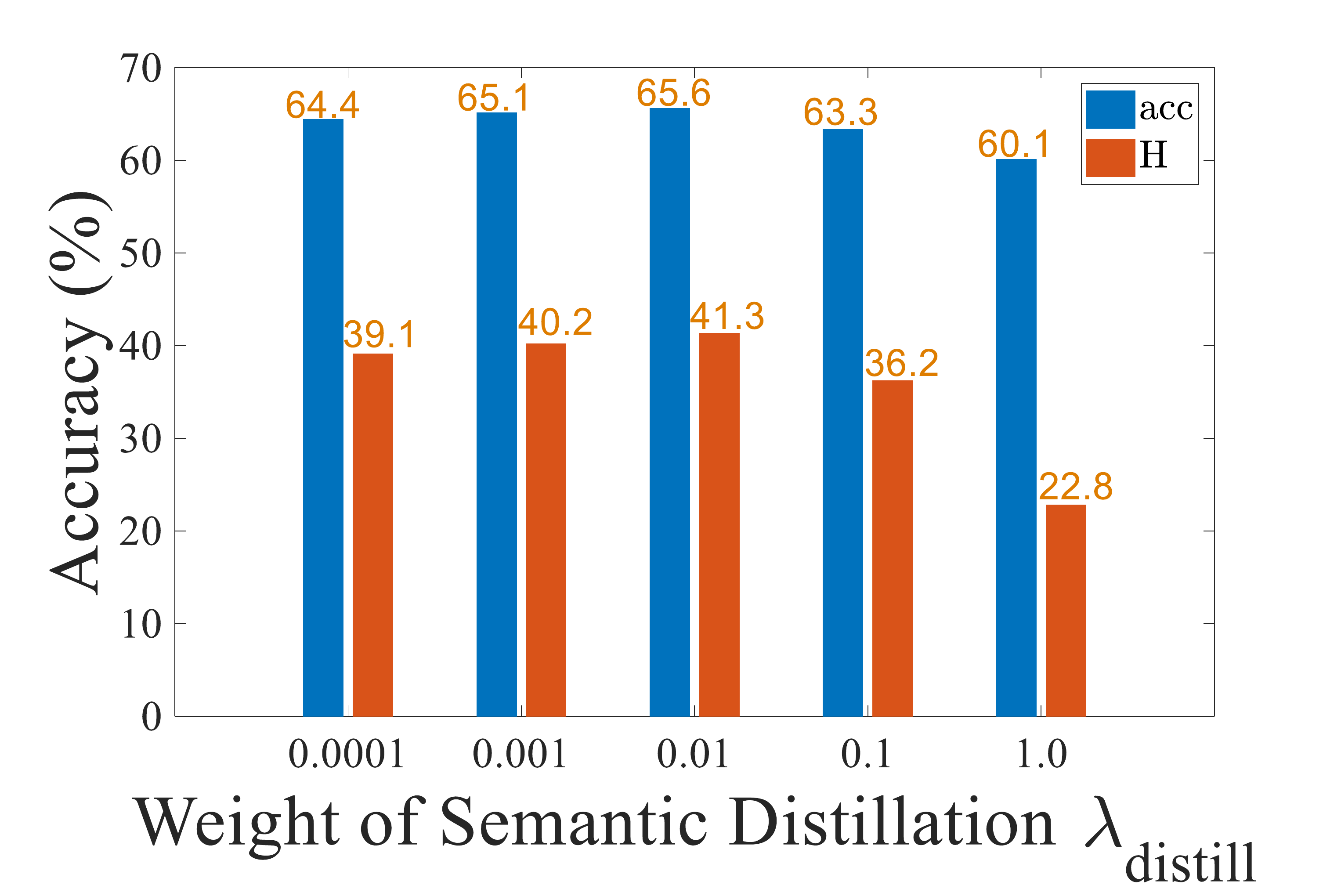}\\ \vspace{-1mm}
			\hspace{-2mm}(a) \hspace{4cm} (b) \hspace{3.5cm} (c) \hspace{3.8cm} (d) \\\vspace{-3mm}
			\caption{The effects of $\lambda_{\text{cal}}$ on (a) CUB and (b) SUN. The effects of $\lambda_{\text{distill}}$ on (c) CUB and (d) SUN. (Best viewed in color)}\vspace{-9mm}
			\label{fig:loss_cal}
		\end{center}
	\end{figure*}

	\subsection{Abaltion Studies}\label{sec4.2}\vspace{-2mm}
	To provide further insight into our MSDN, we conduct ablation studies to evaluate the effectiveness of our MSDN in terms of the V$\rightarrow$A attention sub-net (denoted as MSDN(V$\rightarrow$A) w/o $\mathcal{L}_{\text{distill}}$), A$\rightarrow$V attention sub-net (denoted as MSDN( A$\rightarrow$V) w/o $\mathcal{L}_{\text{distill}}$), semantic distillation loss (\textit{i.e.}, MSDN(V$\rightarrow$A) w/ $\mathcal{L}_{\text{distill}}$, MSDN(A$\rightarrow$V) w/ $\mathcal{L}_{\text{distill}}$), semantic distillation loss with JSD (\textit{i.e.}, MSDN w/ $\mathcal{L}_{\text{distill}}$(JSD)) and $\ell_2$ (\textit{i.e.}, MSDN w/ $\mathcal{L}_{\text{distill}}$($\ell_2$)) . Our results are shown in Table \ref{table:ablation}. Compared to the baseline, MSDN only employs the single attention sub-net without semantic distillation achieving significant improvements. For example,  MSDN(V$\rightarrow$A) w/o $\mathcal{L}_{\text{distill}}$ achieves the gains of acc/H by 8.6\%/6.3\% and 4.4\%/3.3\% on CUB and SUN respectively, MSDN(A$\rightarrow$V) w/o $\mathcal{L}_{\text{distill}}$ gets the acc/H improvements of 16.0\%/16.3\% and 9.0\%/8.0\% on CUB and SUN respectively. This is beneficial from that MSDN refines the visual features, alleviating the cross-dataset bias problem \cite{Chen2021FREE}. If MSDN is optimized by the semantic distillation loss, its results can be further improved, \textit{e.g.},  MSDN(V$\rightarrow$A) improves the Harmonic mean by 5.4\% and 4.8\% on CUB and SUN, respectively. These results show that semantic distillation encourages the two mutual attention sub-nets to learn collaboratively and teach each other, and thus the intrinsic semantic representations can be distilled for knowledge transfer. When the semantic distillation loss only uses one distance, i.e., JSD or $\ell_2$, the distillating capasity of MSDN are limited. Moreover, our full model ensembles the complementary embeddings learned by the two mutual attention sub-nets to further improve the feature representations, achieving acc/Harmonic mean improvements of 18.7\%/19.0\% and 11.0\%/10.8\% on CUB and SUN over the baseline, respectively. 

	\subsection{Qualitative Results}\label{sec4.3}\vspace{-2mm}
	\noindent\textbf{Visualization of Attention Maps.}
	To intuitively show the effectiveness of our MSDN at distilling the intrinsic semantic, we visualize the attention maps learned by the two mutual attention sub-nets, \textit{e.g.}, MSDN(A$\rightarrow$V) and MSDN(V$\rightarrow$A). As shown in Fig. \ref{fig:att}, MSDN(A$\rightarrow$V) and MSDN(V$\rightarrow$A) sub-nets effectively learn the attribute-based visual features and visual-based attribute features for representing the discriminative attribute localizations. MSDN(A$\rightarrow$V) and MSDN(V$\rightarrow$A) can similarly learn the most important semantic representation, which is beneficial from mutual learning for semantic distillation. Furthermore, the two attention sub-nets also learn the complementary attribute feature localization for each other. For example, MSDN(A$\rightarrow$V) can well learn the key semantic of “under tail white” but no “shape chicken-like marsh” for \textit{Carolina Wren}, while  MSDN(V$\rightarrow$A) confidently learn the important semantic “shape chicken-like marsh” but not “under tail white”, Thus, our full MSDN achieves significant performance both in seen and unseen classes.

	\noindent\textbf{t-SNE Visualizations.}
	As shown in Fig. \ref{fig:tsne}, we also present the t-SNE visualization \cite{Maaten2008VisualizingDU} of visual features for seen and unseen classes on CUB, learned by the baseline, MSDN(A$\rightarrow$V), MSDN(V$\rightarrow$A), and combination of the two attention sub-nets (\textit{i.e.}, MSDN(V$\rightarrow$A and A$\rightarrow$V)). Compared to the baseline, our models learn the intrinsic semantic representations both in seen and unseen classes. This shows that our MSDN can simultaneously learn the discriminative and transferable features for effective knowledge transfer in ZSL.  As such, our MSDN achieves significant improvement over baseline.\vspace{-3mm}
	
	\subsection{Hyperparameter Analysis}\label{sec4.4}\vspace{-2mm}
	\noindent\textbf{Effects of Combination Coefficients.}
	we conduct experiments to determine the effectiveness of the combination coefficients ($\alpha_{1},\alpha_{2}$) between attribute$\rightarrow$visual and visual$\rightarrow$attribute attention sub-nets.  As shown in Fig. \ref{fig:combination}, MSDN performs poorly when $\alpha_{1}/\alpha_{2}$ is set too small or large, because both the attribute-based visual features and visual-based attribute features are complementary for discriminative semantic embedding representations. When the combination coefficients $\alpha_{1},\alpha_{2}$ are set to (0.9,0.1) and (0.7,0.3) on CUB and SUN, respectively, MSDN achieves the best results. 
	
	\noindent\textbf{Effects of Loss Weights.} Here we study how to set the related loss weights of MSDN: $\lambda_{\text{cal}}$ and $\lambda_{\text{dstill}}$, which control the self-calibration term and semantic distillation loss, respectively. Based on the results in Fig. \ref{fig:loss_cal} (a) and (b), we choose $\lambda_{\text{cal}}$ as 0.1  for CUB/AWA2. Since the number of seen classes is much larger than the number of unseen classes and per class only contains 16 training images on SUN, thus it usually overfits unseen classes. As such, we set $\lambda_{\text{cal}}$ to 0.0 for SUN. According to the results in Fig. \ref{fig:loss_cal} (c) and (d), we set $\lambda_{\text{dstill}}$ to 0.001 and 0.01 for CUB/AWA2 and SUN, respectively.\vspace{-2mm}

	%

	\section{Conclusion and Discussion}\vspace{-3mm}
	In this paper, we propose a novel mutually semantic distillation network (MSDN) for ZSL. MSDN consists of two mutual attention sub-nets, \textit{i.e.},  attribute$\rightarrow$visual and visual$\rightarrow$semantic attention sub-nets, which learns attribute-based visual features and visual-based attribute features for semantic embedding representations, respectively. To encourage mutual learning between the two attention sub-nets, we introduce a semantic distillation loss that aligns each other's class posterior probabilities. Thus, MSDN distills the intrinsic semantic representations between visual and attribute features for effective knowledge transfer of ZSL. Extensive experiments on three popular benchmarks show the superiority of MSDN. we believe that our work will also facilitate the development of other visual-and-language learning systems, \textit{e.g.}, visual question answering \cite{Agrawal2015VQAVQ}.

	\section*{Acknowledgements} \vspace{-3mm}
	This work is partially supported by NSFC~(61772220, 62172177, 62006244), Special projects for technological innovation in Hubei Province~(2018ACA135), Key R\&D Plan of Hubei Province~(2020BAB027),  Natural Science Foundation of Hubei Province (2021CFB332), Young Elite Scientist Sponsorship Program of China Association for Science and Technology (YESS20200140).
	
	{\small
		\bibliographystyle{ieee_fullname}
		\bibliography{mybibfile}
	}
	
\end{document}